\documentclass[runningheads]{llncs}
\usepackage{graphicx}
\usepackage{amsmath,amssymb} 
\usepackage{color}

\usepackage{array}
\usepackage{makecell}

\newcommand{\COG}{{\tt COG} }
\newcommand{\COGns}{{\tt COG}}

\begin{document}

\title{A Dataset and Architecture for Visual Reasoning with a Working Memory} 

\titlerunning{Visual Reasoning with Working Memory}

\authorrunning{G.R. Yang, I. Ganichev, X.-J. Wang, J. Shlens, D. Sussillo}


\author{Guangyu Robert Yang\textsuperscript{1,$\dagger$,$\ddagger$,*} \and Igor Ganichev\textsuperscript{2,*} \and Xiao-Jing Wang\textsuperscript{1} \and Jonathon Shlens\textsuperscript{2} \and David Sussillo\textsuperscript{2}}


\institute{\textsuperscript{1} Center for Neural Science, New York University \\\textsuperscript{2} Google Brain \\\textsuperscript{$\dagger$} Work done as an intern at Google Brain \\\textsuperscript{$\ddagger$} Present address: Department of Neuroscience, Columbia University \\\textsuperscript{*} equal contribution\\
\email{ robert.yang@columbia.edu,iga@google.com,xjwang@nyu.edu,\\
\{shlens,sussillo\}@google.com}}

\maketitle

\begin{abstract}
A vexing problem in artificial intelligence is reasoning about events that occur in complex, changing visual stimuli such as in video analysis or game play. Inspired by a rich tradition of visual reasoning and memory in cognitive psychology and neuroscience, we developed an artificial, configurable visual question and answer dataset (\COGns) to parallel experiments in humans and animals. \COG is much simpler than the general problem of video analysis, yet it addresses many of the problems relating to visual and logical reasoning and memory -- problems that remain challenging for modern deep learning architectures. We additionally propose a deep learning architecture that performs competitively on other diagnostic VQA datasets (i.e. CLEVR) as well as easy settings of the \COG dataset. However, several settings of \COG result in datasets that are progressively more challenging to learn. After training, the network can zero-shot generalize to many new tasks. Preliminary analyses of the network architectures trained on \COG demonstrate that the network accomplishes the task in a manner interpretable to humans.

\end{abstract}

\section{Introduction}

\begin{figure}[t]
\begin{center}
\centerline{\includegraphics[width=1.0\linewidth]{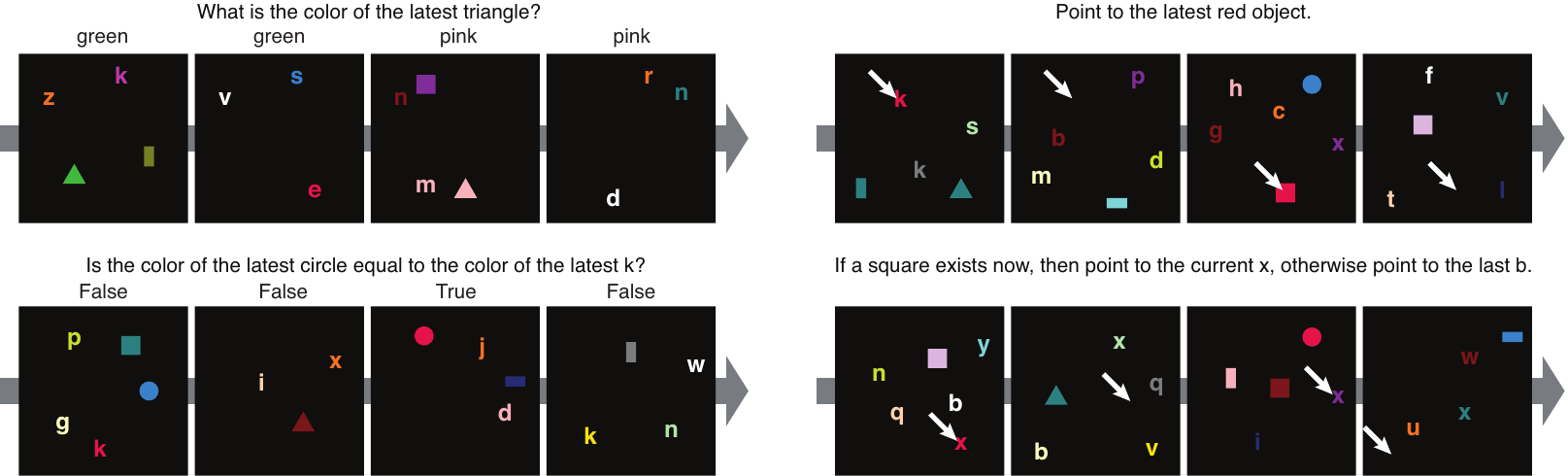}}
\caption{Sample sequence of images and instruction from the \COG dataset. Tasks in the \COG dataset test aspects of object recognition, relational understanding and the manipulation and adaptation of memory to address a problem. Each task can involve objects shown in the current image and in previous images. Note that in the final example, the instruction involves the \textit{last} instead of the \textit{latest} ``b''. The former excludes the current ``b'' in the image. Target pointing response for each image is shown (white arrow).  High-resolution image and proper English are used for clarity.}
\label{fig:illustration}
\end{center}
\vskip -0.4in
\end{figure}

A major goal of artificial intelligence is to build systems that powerfully and flexibly reason about the sensory environment \cite{hassabis2017neuroscience} \footnote{Published at European Conference on Computer Vision (ECCV) 2018.}.
Vision provides an extremely rich and highly applicable domain for exercising our ability to build systems that form logical inferences on complex stimuli \cite{hu2017learning,johnson2017inferring,santoro2017simple,perez17}. 
One avenue for studying visual reasoning has been Visual Question Answering (VQA) datasets where a model learns to correctly answer challenging natural language questions about static images \cite{antol2015vqa,gao2015you,malinowski2014multi,zhu2016visual7w}.
While advances on these multi-modal datasets have been significant, these datasets highlight several limitations to current approaches. First, it is uncertain the degree to which models trained on VQA datasets merely follow statistical cues inherent in the images, instead of reasoning about the logical components of a problem \cite{johnson17,sturm2014simple,agrawal2016analyzing,winograd1972}.
Second, such datasets avoid the complications of time and memory -- both integral factors in the design of intelligent agents \cite{hassabis2017neuroscience,mnih2013playing,mnih2015human,vinyals2017starcraft} and the analysis and summarization of videos \cite{KarpathyCVPR14,abu2016youtube,caba2015activitynet}.

To address the shortcomings related to logical reasoning about spatial relationships in VQA datasets, Johnson and colleagues \cite{johnson17} recently proposed CLEVR to directly test models for elementary visual reasoning, to be used in conjunction with other VQA datasets (e.g. \cite{antol2015vqa,gao2015you,malinowski2014multi,zhu2016visual7w}). The CLEVR dataset provides artificial, static images and natural language questions about those images that exercise the ability of a model to perform logical and visual reasoning. Recent work has demonstrated networks that achieve impressive performance with near perfect accuracy \cite{perez17,santoro2017simple,arad2018compositional}.

In this work, we address the second limitation concerning time and memory in visual reasoning.  A reasoning agent must remember relevant pieces of its visual history, ignore irrelevant detail, update and manipulate a memory based on new information, and exploit this memory at later times to make decisions.  Our approach is to create an artificial dataset that has many of the complexities found in temporally varying data, yet also to eschew much of the visual complexity and technical difficulty of working with video (e.g. video decoding, redundancy across temporally-smooth frames).  In particular, we take inspiration from decades of research in cognitive psychology \cite{diamond2013executive,miyake2000unity,berg1948simple,milner1963effects,baddeley1992working} and modern systems neuroscience (e.g.  \cite{miller1996neural,miller2001integrative,newsome1989neuronal,romo2003cognitive,mante2013context,rigotti2013importance}) -- fields which have a long history of dissecting visual reasoning into core components based on spatial and logical reasoning, memory compositionality, and semantic understanding. Towards this end, we build an artificial dataset -- termed \COG -- that exercises visual reasoning in time, in parallel with human cognitive experiments \cite{yntema1963keeping,zelazo1996age,owen2005n}.

The \COG dataset is based on a programmatic language that builds a battery of task triplets: an image sequence, a verbal instruction, and a sequence of correct answers. These randomly generated triplets exercise visual reasoning across a large array of tasks and require semantic comprehension of text, visual perception of each image in the sequence, and a working memory to determine the temporally varying answers (Figure \ref{fig:illustration}). We highlight several parameters in the programmatic language that allow researchers to modulate the problem difficulty from easy to challenging settings.

Finally, we introduce a multi-modal recurrent architecture for visual reasoning with memory. This network combines semantic and visual modules with a stateful {\it controller} that modulates visual attention and memory in order to correctly perform a visual task.  We demonstrate that this model achieves near state-of-the-art performance on the CLEVR dataset.  In addition, this network provides a strong baseline that achieves good performance on the \COG dataset across an array of settings. Through ablation studies and an analysis of network dynamics, we find that the network employs human-interpretable, attention mechanisms to solve these visual reasoning tasks. We hope that the \COG dataset, corresponding architecture, and associated baseline provide a helpful benchmark for studying reasoning in time-varying visual stimuli \footnote{The \COG dataset and code for the network architecture are open-sourced at https://github.com/google/cog.}.

\section{Related Work}

It is broadly understood in the AI community that memory is a largely unsolved problem and there are many efforts underway to understand this problem, e.g. studied in \cite{DBLP:journals/corr/GravesWD14,DBLP:journals/corr/JoulinM15,collins2017capacity}.  The ability of sequential models to compute in time is notably limited by memory horizon and memory capacity \cite{collins2017capacity} as measured in synthetic sequential datasets \cite{hochreiter97}. Indeed, a large constraint in training network models to perform generic Turing-complete operations is the ability to train systems that compute in time \cite{DBLP:journals/nature/GravesWRHDGCGRA16,collins2017capacity}.

Developing computer systems that comprehend time-varying sequence of images is a prominent interest in video understanding \cite{abu2016youtube,caba2015activitynet,kay2017kinetics} and intelligent video game agents \cite{mnih2013playing,mnih2015human,hassabis2017neuroscience}. While some attempts have used a feed-forward architecture (e.g. \cite{mnih2013playing}, baseline model in \cite{vinyals2017starcraft}), much work has been invested in building video analysis and game agents that contain a memory component \cite{vinyals2017starcraft,ng2015beyond}. These types of systems are often limited by the flexibility of network memory systems, and it is not clear the degree to which these systems reason based on complex relationships from past visual imagery.  

Let us consider Visual Question Answering (VQA) datasets based on single, static images \cite{antol2015vqa,gao2015you,malinowski2014multi,zhu2016visual7w}. These datasets construct natural language questions to probe the logical understanding of a network about natural images. There has been strong suggestion in the literature that networks trained on these datasets focus on statistical regularities for the prediction tasks, whereby a system may ``cheat'' to superficially solve a given task \cite{sturm2014simple,johnson17}. Towards that end, several researchers proposed to build an auxiliary diagnostic, synthetic datasets to uncover these potential failure modes and highlight logical comprehension (e.g. attribute identification, counting, comparison, multiple attention, and logical operations) \cite{johnson17,weston2015towards,zitnick2013bringing,winograd1972,kuhnle2017shapeworld}. Further, many specialized neural network architectures focused on multi-task learning have been proposed to address this problem by leveraging attention \cite{xu2016ask}, external memory \cite{DBLP:journals/corr/GravesWD14,DBLP:journals/corr/JoulinM15}, a family of feature-wise transformations \cite{Dumoulin2017,perez17}, explicitly parsing a task into executable sub-tasks \cite{johnson2017inferring,hu2017learning}, and inferring relations between pairs of objects \cite{santoro2017simple}.

Our contribution takes direct inspiration from this previous work on single images but focuses on the aspects of time and memory. A second source of inspiration is the long line of cognitive neuroscience literature that has focused on developing a battery of sequential visual tasks to exercise and measure specific attributes of visual working memory \cite{diamond2013executive,luck1997capacity,miller1996neural}. Several lines of cognitive psychology and neuroscience have developed multitudes of visual tasks in time that exercise attribute identification, counting, comparison, multiple attention, and logical operations \cite{yntema1963keeping,miller1996neural,zelazo1996age,owen2005n,newsome1989neuronal,romo2003cognitive,mante2013context,rigotti2013importance} (see references therein). This work emphasizes compositionality in task generation -- a key ingredient in generalizing to unseen tasks \cite{cole2013rapid}. Importantly, this literature provides measurements in humans and animals on these tasks as well as discusses the biological circuits and computations that may underlie and explain the variability in performance\cite{miller2001integrative,newsome1989neuronal,romo2003cognitive,mante2013context,rigotti2013importance}.

\section{The \COG dataset}
\label{sec:cog}

We designed a large set of tasks that requires a broad range of cognitive skills to solve, especially working memory. One major goal of this dataset is to build a compositional set of tasks that include variants of many cognitive tasks studied in humans and other animals  \cite{yntema1963keeping,miller1996neural,zelazo1996age,owen2005n,newsome1989neuronal,romo2003cognitive,mante2013context,rigotti2013importance} (see also Introduction and Related Work).

The dataset contains triplets of a task instruction, sequences of synthetic images, and sequences of target responses (see Figure \ref{fig:illustration} for examples). Each image consists of a number of simple objects that vary in color, shape, and location. There are 19 possible colors and 33 possible shapes (6 geometric shapes and 26 lower-case English letters). The network needs to generate a verbal or pointing response for every image.

\begin{figure}[ht]
\begin{center}
\centerline{\includegraphics[width=1.0\linewidth]{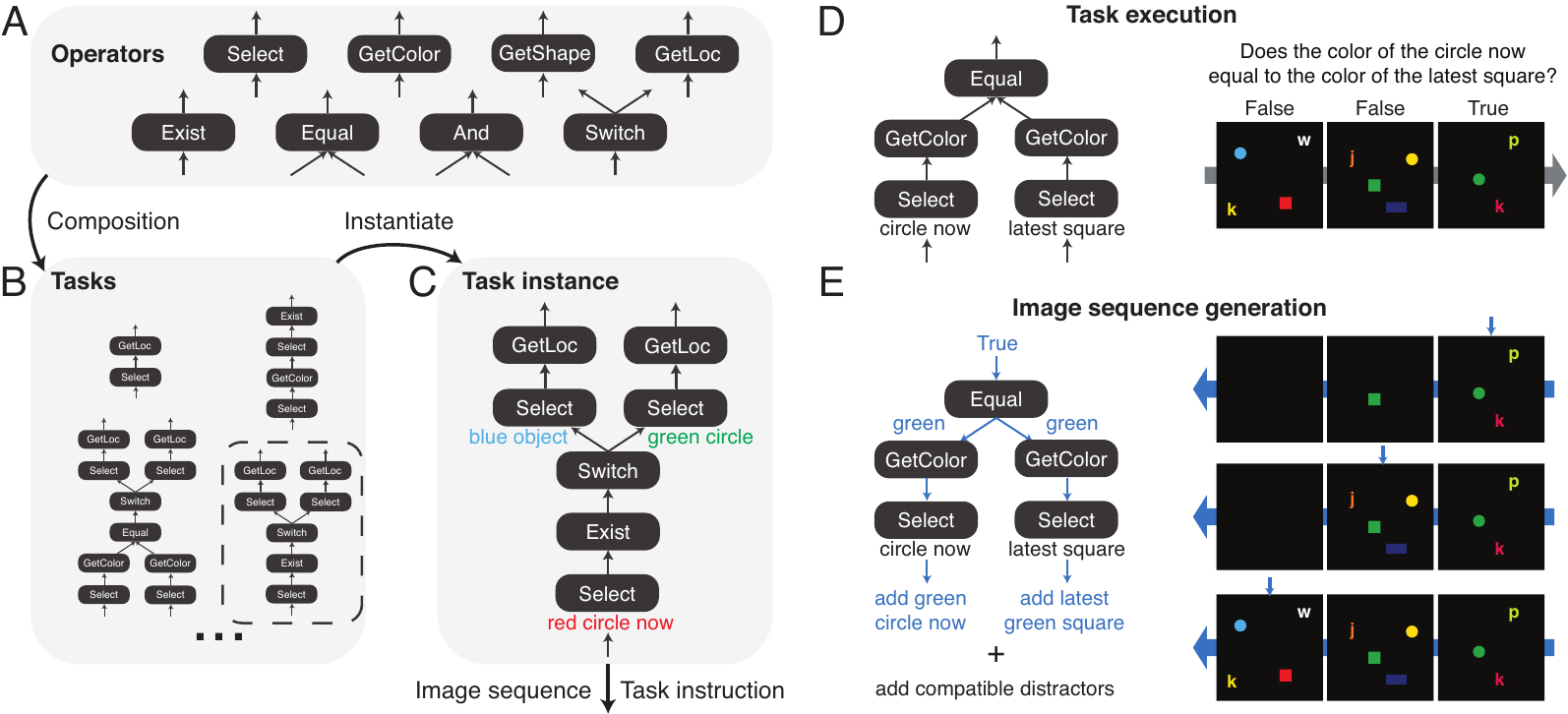}}
\caption{Generating the compositional \COG dataset. The \COG dataset is based on a set of operators (\textbf{A}), which are combined to form various task graphs (\textbf{B}). (\textbf{C}) A task is instantiated by specifying the attributes of all operators in its graph. A task instance is used to generate both the image sequence and the semantic task instruction. (\textbf{D}) Forward pass through the graph and the image sequence for normal task execution. (\textbf{E}) Generating a consistent, minimally biased image sequence requires a backward pass through the graph in a reverse topological order and through the image sequence in the reverse chronological order.}
\label{datasetgen_scheme}
\end{center}
\vskip -0.2in
\end{figure}

To build a large set of tasks, we first describe all potential tasks using a common, unified framework. Each task in the dataset is defined abstractly and constructed compositionally from basic building blocks, namely {\it operators}. An operator performs a basic computation, such as selecting an object based on attributes (color, shape, etc.) or comparing two attributes (Figure \ref{datasetgen_scheme}A). The operators are defined abstractly without specifying the exact attributes involved. A task is formed by a directed acyclic graph of operators (Figure \ref{datasetgen_scheme}B). Finally, we instantiate a task by specifying all relevant attributes in its graph (Figure \ref{datasetgen_scheme}C). The task instance is used to generate both the verbal task instruction and minimally-biased image sequences. Many image sequences can be generated from the same task instance.

There are 8 operators, 44 tasks, and more than 2 trillion possible task instances in the dataset (see Appendix for more sample task instances). We vary the number of images ($F$), the maximum memory duration ($M_{\mathrm{max}}$), and the maximum number of distractors on each image ($D_{\mathrm{max}}$) to explore the memory and capacity of our proposed model and systematically vary the task difficulty. When not explicitly stated, we use a canonical setting with $F=4$, $M_{\mathrm{max}}=3$, and $D_{\mathrm{max}}=1$ (see Appendix for the rationale).

The \COG dataset is in many ways similar to the CLEVR dataset \cite{johnson17}. Both contain synthetic visual inputs and tasks defined as operator graphs (functional programs). However, \COG differs from CLEVR in two important ways. First, all tasks in the \COG dataset can involve objects shown in the past, due to the sequential nature of their inputs.
Second, in the \COG dataset, visual inputs with minimal response bias can be generated on the fly.

An operator is a simple function that receives and produces abstract data types such as an attribute, an object, a set of objects, a spatial range, or a Boolean. There are 8 operators in total: \textit{Select}, \textit{GetColor}, \textit{GetShape}, \textit{GetLoc}, \textit{Exist}, \textit{Equal}, \textit{And}, and \textit{Switch} (see Appendix for details). Using these 8 operators, the \COG dataset currently contains 44 tasks, with the number of operators in each task graph ranging from 2 to 11. Each task instruction is obtained from a task instance by traversing the task graph and combining pieces of text associated with each operator. It is straightforward to extend the \COG dataset by introducing new operators.

Response bias is a major concern when designing a synthetic dataset. Neural networks may achieve high accuracy in a dataset by exploiting its bias. Rejection sampling can be used to ensure an \textit{ad hoc} balanced response distribution \cite{johnson17}. We developed a method for the \COG dataset to generate minimally-biased synthetic image sequences tailored to individual tasks. 

In short, we first determine the minimally-biased responses (target outputs), then we generate images (inputs) that would lead to these specified responses. The images are generated in the reversed order of normal task execution (Figure \ref{datasetgen_scheme}D, E). During generation, images are visited in the reverse chronological order and the task graph traversed in a reverse topological order (Figure \ref{datasetgen_scheme}E). When visiting an operator, if its target output is not already specified, we randomly choose one from all allowable outputs. Based on the specified output, the image is modified accordingly and/or the supposed input is passed on to the next operator(s) as their target outputs (see details in Appendix). In addition, we can place a uniformly-distributed $D \sim U(1, D_{\mathrm{max}})$ distractors, then delete those that interfere with the normal task execution.

\section{The network}
\subsection{General network setup}
\begin{figure}[ht]
\begin{center}
\centerline{\includegraphics[width=1.0\linewidth]{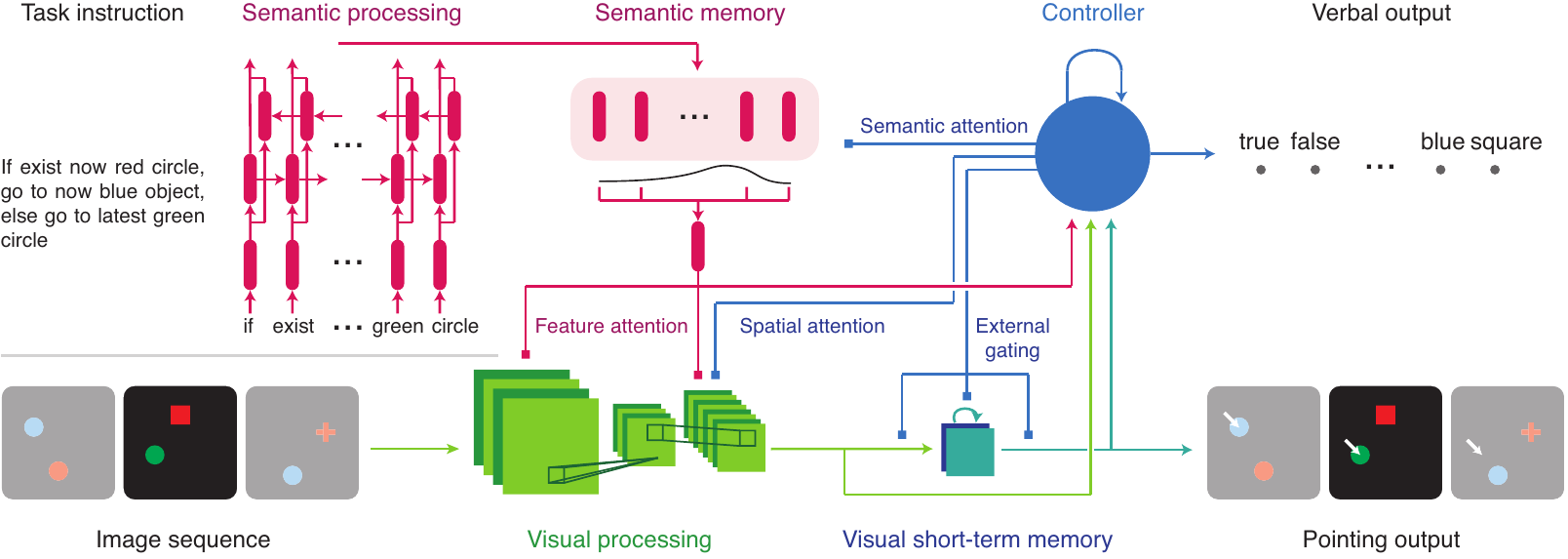}}
\caption{Diagram of the proposed network. A sequence of images are provided as input into a convolutional neural network (green). An instruction in the form of English text is provided into a sequential embedding network (red). A visual short-term memory (vSTM) network holds visual-spatial information in time and provides the pointing output (teal). The vSTM module can be considered a convolutional LSTM network with external gating. A stateful controller (blue) provides all attention and gating signals directly or indirectly. The output of the network is either discrete (verbal) or 2D continuous (pointing).}
\label{fig:network_scheme}
\end{center}
\vskip -0.2in
\end{figure}

Overall, the network contains four major systems (Figure \ref{fig:network_scheme}). The visual system processes the images. The semantic system processes the task instructions. The visual short-term memory system maintains the processed visual information, and provides outputs that guide the pointing response. Finally, the control system integrates converging information from all other systems, uses several attention and gating mechanisms to regulate how other systems process inputs and generate outputs, and provides verbal outputs. Critically, the network is allowed multiple time steps to ``ponder'' about each image \cite{graves2016adaptive}, giving it the potential to solve multi-step reasoning problems naturally through iteration.

\subsection{Visual processing system}
The visual system processes the raw input images. The visual inputs are $112\times 112$ images and are processed by 4 convolutional layers with 32, 64, 64, 128 feature maps respectively. Each convolutional layer employs $3\times 3$ kernels and is followed by a $2\times 2$ max-pooling layer, batch-normalization \cite{ioffe15}, and a rectified-linear activation function. This simple and relatively shallow architecture was shown to be sufficient for the CLEVR dataset \cite{johnson17,santoro2017simple}.

The last two layers of the convolutional network are subject to feature and spatial attention. Feature attention scales and shifts the batch normalization parameters of
individual feature maps, such that the activity of all neurons within a feature map are multiplied and added by two scalars.
This particular implementation of feature attention has been termed conditional batch-normalization or feature-wise linear modulation (FiLM) \cite{Dumoulin2017,perez17}. FiLM is a critical component for the model that achieved near state-of-the-art performance on the CLEVR dataset \cite{perez17}. Soft spatial attention \cite{xu15} is applied to the top convolutional layer following feature attention and the activation function. It multiplies the activity of all neurons with the same spatial preferences using a positive scalar.

\subsection{Semantic processing system}
The semantic processing system receives a task instruction and generates a semantic memory that the controller can later attend to. Conceptually, it produces a semantic memory -- a contextualized representation of each word in the instruction -- before the task is actually being performed. At each pondering step when performing the task, the controller can attend to individual parts of the semantic memory corresponding to different words or phrases.

Each word is mapped to a 64-dimensional trainable embedding vector, then sequentially fed into an 128-unit bidirectional Long Short-Term Memory (LSTM) network \cite{schuster97,hochreiter97}. The outputs of the bidirectional LSTM for all words form a semantic memory of size $(n_{\mathrm{word}}, n_{\mathrm{rule}}^{\mathrm{(out)}})$, where $n_{\mathrm{word}}$ is the number of words in the instruction, and $n_{\mathrm{rule}}^{\mathrm{(out)}}=128$ is the dimension of the output vector.

Each $n_{\mathrm{rule}}^{\mathrm{(out)}}$-dimensional vector in the semantic memory forms a key. For semantic attention, a query vector of the same dimension $n_{\mathrm{rule}}^{\mathrm{(out)}}$ is used to retrieve the semantic memory by summing up all the keys weighted by their similarities to the query. We used Bahdanau attention \cite{bahdanau14}, which computes the similarity between the query $\mathbf{q}$ and a key $\mathbf{k}$ as $\sum_{i=1}^{n_{\mathrm{rule}}^{\mathrm{(out)}}} v_i \cdot \mathrm{tanh}(q_i+k_i)$, where $\mathbf{v}$ is trained.

\subsection{Visual short-term memory system}
To utilize the spatial information preserved in the visual system for the pointing output, the top layer of the convolutional network feeds into a visual short-term memory module, which in turn projects to a group of pointing output neurons. This structure is also inspired by the posterior parietal cortex in the brain that maintains visual-spatial information to guide action \cite{andersen1997multimodal}.

The visual short-term memory (vSTM) module is an extension of a 2-d convolutional LSTM network \cite{xingjian2015convolutional} in which the gating mechanisms are conditioned on external information. The vSTM module consists of a number of 2-D feature maps, while the input and output connections are both convolutional. There is currently no recurrent connections within the vSTM module besides the forget gate. The state $c_t$ and output $h_t$ of this module at step $t$ are
\begin{eqnarray}
    c_t &=& f_t * c_{t-1} + i_t * x_t, \\
    h_t &=& o_t * \mathrm{tanh}(c_t),
\end{eqnarray}
where * indicates a convolution. This vSTM module differs from a convolutional LSTM network mainly in that the input $i_t$, forget $f_t$, and output gates $o_t$ are not self-generated. Instead, they are all provided externally from the controller. In addition, the input $x_t$ is not directly fed into the network, but a convolutional layer can be applied in between.

All convolutions are currently set to be $1 \times 1$. Equivalently, each feature map of the vSTM module adds its gated previous activity with a weighted combination of the post-attention activity of all feature maps from the top layer of the visual system. Finally, the activity of all vSTM feature maps is combined to generate a single spatial output map $h_t$.

\subsection{Controller}
To synthesize information across the entire network, we include a controller that receives feedforward inputs from all other systems and generates feedback attention and gating signals. This architecture is further inspired by the prefrontal cortex of the brain \cite{miller2001integrative}. The controller is a Gated Recurrent Unit (GRU) network. At each pondering step, the post-attention activity of the top visual layer is processed through a 128-unit fully connected layer, concatenated with the retrieved semantic memory and the vSTM module output, then fed into the controller. In addition, the activity of the top visual layer is summed up across space and provided to the controller.

The controller generates queries for the semantic memory through a linear feedforward network. The retrieved semantic memory then generates the feature attention through another linear feedforward network. The controller generates the 49-dimensional soft spatial attention through a two layer feedforward network, with a 10-unit hidden layer and a rectified-linear activation function, followed by a softmax normalization. Finally, the controller state is concatenated with the retrieved semantic memory to generate the input, forget, and output gates used in the vSTM module through a linear feedforward network followed by a sigmoidal activation function.

\subsection{Output, loss, and optimization}
The verbal output is a single word, and the pointing output is the $(x,y)$ coordinates of pointing. Each coordinate is between 0 and 1. A loss function is defined for each output, and only one loss function is used for every task. The verbal output uses a cross-entropy loss. To ensure the pointing output loss is comparable in scale to the verbal output loss, we include a group of pointing output neurons on a $7 \times 7$ spatial grid, and compute a cross-entropy loss over this group of neurons. Given a target $(x,y)$ coordinates, we use a Gaussian distribution centered at the target location with $\sigma=0.1$ as the target probability distribution of the pointing output neurons.

For each image, the loss is based on the output at the last pondering step. No loss is used if there is no valid output for a given image. We use a L2 regularization of strength 2e-5 on all the weights. We clip the gradient norm at $10$ for \COG and at $80$ for CLEVR. We clip the controller state norm at $10000$ for \COG and $5000$ for CLEVR. We also trained all initial states of the recurrent networks. The network is trained end-to-end with Adam \cite{kingma14}, combined with a learning rate decay schedule. 

\section{Results}

\subsection{Intuitive and interpretable solutions on the CLEVR dataset}

To demonstrate the reasoning capability of our proposed network, we trained it on the CLEVR dataset \cite{johnson17}, even though there is no explicit need for working memory in CLEVR. The network achieved an overall test accuracy of 96.8\% on CLEVR, surpassing human-level performance and comparable with other state-of-the-art methods \cite{santoro2017simple,perez17,arad2018compositional} (Table \ref{table:clevr_accuracy}, see Appendix for more details).

Images were first resized to $128 \times 128$, then randomly cropped or resized to $112 \times 112$ during training and validation/testing respectively. In the best-performing network, the controller used 12 pondering steps per image. Feature attention was applied to the top two convolutional layers. The vSTM module was disabled since there is no pointing output.

\setlength{\tabcolsep}{4pt}
\begin{table}
\begin{center}
\resizebox{\textwidth}{!}{%
\begin{tabular}{l|m{0.1\linewidth}m{0.1\linewidth}m{0.1\linewidth}m{0.1\linewidth}m{0.1\linewidth}m{0.1\linewidth}}
\hline
Model & Overall & Count & Exist & Compare\newline Numbers & Query\newline Attribute & Compare\newline Attribute \\
\hline
Human \cite{johnson17}    & 92.6 & 86.7 & 96.6 & 86.5 & 95.0 & 96.0  \\
\hline
Q-type baseline \cite{johnson17} & 41.8 & 34.6 & 50.2 & 51.0 & 36.0 & 51.3\\
CNN+LSTM+SA \cite{santoro2017simple} & 76.6 & 64.4 & 82.7 & 77.4 & 82.6 & 75.4 \\
CNN+LSTM+RN \cite{santoro2017simple}    & 95.5 & 90.1 & 97.8 & 93.6 & 97.9 & 97.1         \\
CNN+GRU+FiLM \cite{perez17}     & 97.6 & 94.3 & 99.3 & 93.4 & 99.3 & 99.3\\
MAC* \cite{arad2018compositional} & 98.9 & 97.2 & 99.5 & 99.4 & 99.3 & 99.5 \\
\hline
Our model   & 96.8 & 91.7 & 99.0 & 95.5 & 98.5 & 98.8 \\
\hline
\end{tabular}
}
\end{center}
\caption{CLEVR test accuracies for human, baseline, and top-performing models that relied only on pixel inputs and task instructions during training. (*) denotes use of pretrained models.}
\label{table:clevr_accuracy}
\end{table}
\setlength{\tabcolsep}{1.4pt}

\begin{figure}[ht]
\begin{center}
\centerline{\includegraphics[width=0.9\linewidth]{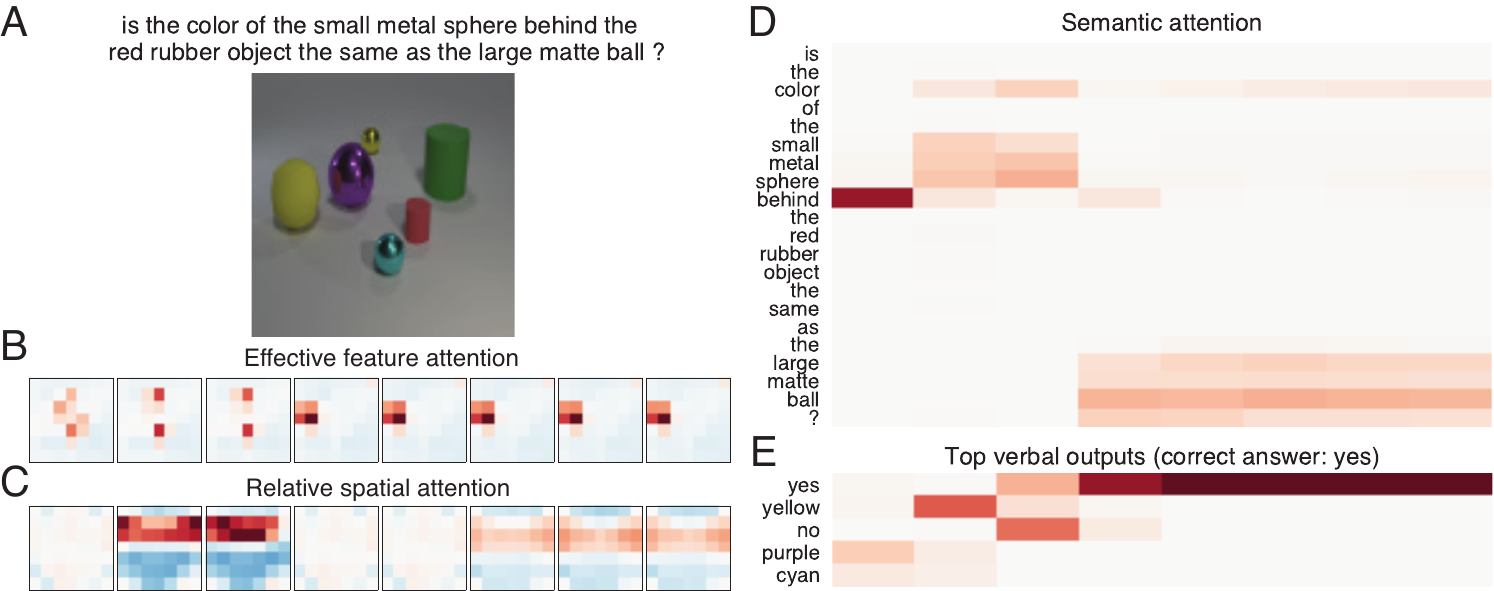}}
\caption{Pondering process of the proposed network, visualized through attention and output for a single CLEVR example. (\textbf{A}) The example question and image from the CLEVR validation set. (\textbf{B}) The effective feature attention map for each pondering step. (\textbf{C}) The relative spatial attention maps. (\textbf{D}) The semantic attention. (\textbf{E}) Top five verbal outputs. Red and blue indicate stronger and weaker, respectively. After simultaneous feature attention to the ``small metal spheres'' and spatial attention to ``behind the red rubber object'', the color of the attended object (yellow) was reflected in the verbal output. Later in the pondering process, the network paid feature attention to the ``large matte ball'', while the correct answer (yes) emerged in the verbal output.}
\label{clevr_attention}
\end{center}
\vskip -0.2in
\end{figure}

The output of the network is human-interpretable and intuitive.
In Figure \ref{clevr_attention}, we illustrate how the verbal output and various attention signals evolved through pondering steps for an example image-question pair. The network answered a long question by decomposing it into small, executable steps. Even though training only relies on verbal outputs at the last pondering steps, the network learned to produce interpretable verbal outputs that reflect its reasoning process.

In Figure \ref{clevr_attention}, we computed effective feature attention as the difference between the normalized activity maps with or without feature attention. To get the post- (or pre-) feature-attention normalized activity map, we average the activity across all feature maps after (or without) feature attention, then divide the activity by its mean. The relative spatial attention is normalized by subtracting the time-averaged spatial attention map. This example network uses 8 pondering steps.

\subsection{Training on the \COG dataset}
Our proposed model achieved a maximum overall test accuracy of 93.7\% on the \COG dataset in the canonical setting (see Section \ref{sec:cog}). In the Appendix, we discuss potential strategies for measuring human accuracy on the \COG dataset. We noticed a small but significant variability in the final accuracy even for networks with the same hyperparameters (mean $\pm$ std: $90.6 \pm 2.8\%$, 50 networks). We found that tasks containing more operators tend to take substantially longer to be learned or remain at lower accuracy (see Appendix for more results). We tried many approaches of reducing variance including various curriculum learning regimes, different weight and bias initializations, different optimizers and their hyperparameters. All approaches we tried either did not significantly reduce the variance or degraded performance.

The best network uses 5 pondering steps for each image. Feature attention is applied to the top layer of the visual network. The vSTM module contains 4 feature maps.

\subsection{Assessing the contribution of model parts through ablation}
The model we proposed contains multiple attention mechanisms, a short-term memory module, and multiple pondering steps. To assess the contribution of each component to the overall accuracy, we trained versions of the network on the CLEVR and the \COG dataset in which one component was ablated from the full network. We also trained a baseline network with all components ablated. The baseline network still contains a CNN for visual processing, a LSTM network for semantic processing, and a GRU network as the controller. To give each ablated network a fair chance, we re-tuned their hyperparameters, with the total number of parameters limited at $110\%$ of the original network, and reported the maximum accuracy.

\begin{figure}[htbp]
\begin{center}
\centerline{\includegraphics[width=1.0\linewidth]{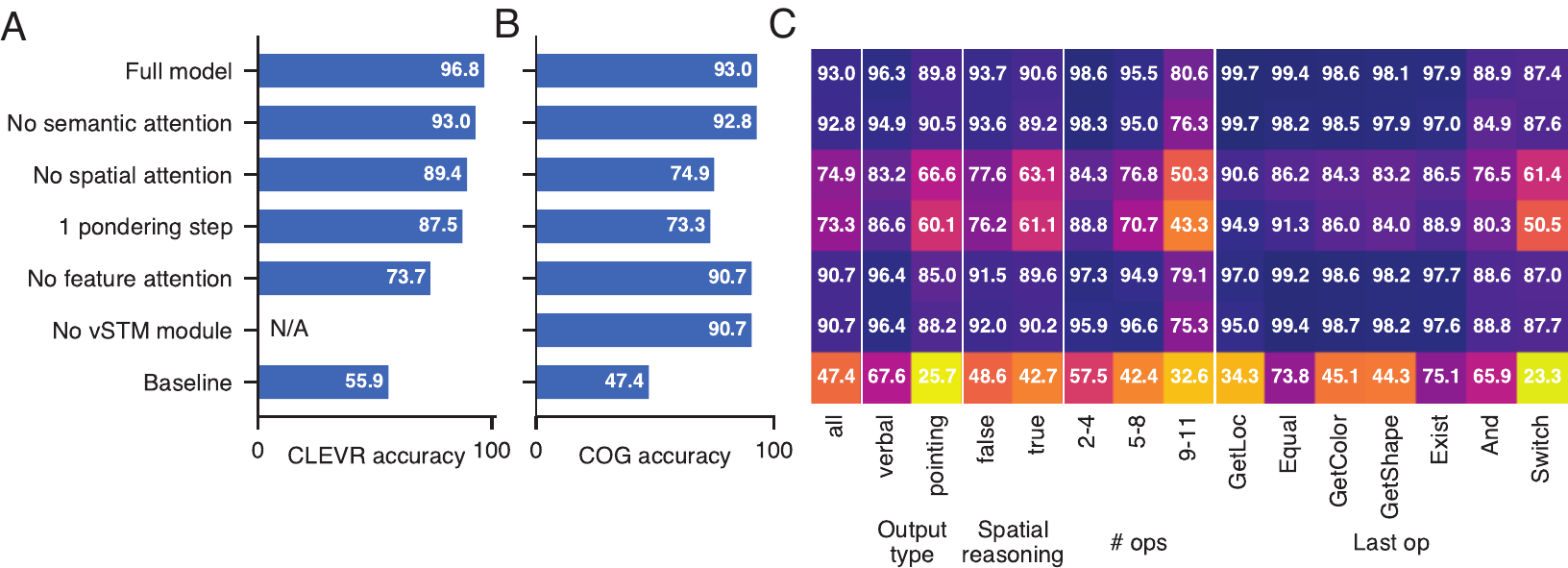}}
\caption{Ablation studies. Overall accuracies for various ablation models on the CLEVR test set (\textbf{A}) and \COG (\textbf{B}). vSTM module is not included in any model for CLEVR. (\textbf{C}) Breaking the \COG accuracies down based on the output type, whether spatial reasoning is involved, the number of operators, and the last operator in the task graph.}
\label{ablation}
\end{center}
\vskip -0.2in
\end{figure}

We found that the baseline network performed poorly on both datasets (Figure \ref{ablation}A, B). To our surprise, the network relies on a different combination of mechanisms to solve the CLEVR and the \COG dataset. The network depends strongly on feature attention for CLEVR (Figure \ref{ablation}A), while it depends strongly on spatial attention for the \COG dataset (Figure \ref{ablation}B). One possible explanation is that there are fewer possible objects in CLEVR (96 combinations compared to 608 combinations in \COGns), making feature attention on $\sim 100$ feature maps better suited to select objects in CLEVR. Having multiple pondering steps is important for both datasets, demonstrating that it is beneficial to solve multi-step reasoning problems through iteration. Although semantic attention has a rather minor impact on the overall accuracy of both datasets, it is more useful for tasks with more operators and longer task instructions (Figure \ref{ablation}C).

\subsection{Exploring the range of difficulty of the \COG dataset}
\begin{figure}[htbp]
\begin{center}
\centerline{\includegraphics[width=0.9\linewidth]{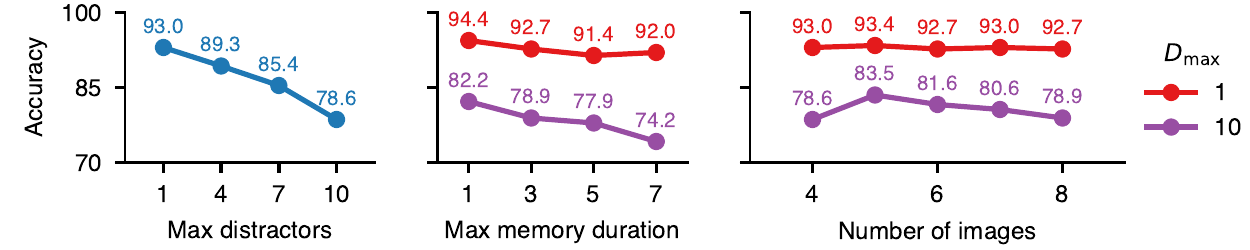}}
\caption{Accuracies on variants of the \COG dataset. From left to right, varying the maximum number of distractors ($D_{\mathrm{max}}$), the maximum memory duration ($M_{\mathrm{max}}$), and the number of images in each sequence ($F$).}
\label{cog_difficult}
\end{center}
\vskip -0.2in
\end{figure}

To explore the range of difficulty in visual reasoning in our dataset, we varied the maximum number of distractors on each image ($D_{\mathrm{max}}$), the maximum memory duration ($M_{\mathrm{max}}$), and the number of images in each sequence ($F$) (Figure \ref{cog_difficult}). For each setting we selected the best network across 50-80 hyper-parameter settings involving model capacity and learning rate schedules. Out of all models explored, the accuracy of the best network drops substantially with more distractors. When there is a large number of distractors, the network accuracy also drops with longer memory duration. These results suggest that the network has difficulty filtering out many distractors and maintaining memory at the same time. However, doubling the number of images does not have a clear effect on the accuracy, which indicates that the network developed a solution that is invariant to the number of images used in the sequence. The harder setting of the \COG dataset with $F=8$, $D_{\mathrm{max}}=10$ and $M_{\mathrm{max}}=7$ can potentially serve as a benchmark for more powerful neural network models.

\subsection{Zero-shot generalization to new tasks}
A hallmark of intelligence is the flexibility and capability to generalize to unseen situations. During training and testing, each image sequence is generated anew, therefore the network is able to generalize to unseen input images. On top of that, the network can generalize to trillions of task instances (new task instructions), although only millions of them are used during training.

\begin{figure}[t]
\begin{center}
\centerline{\includegraphics[width=1.0\linewidth]{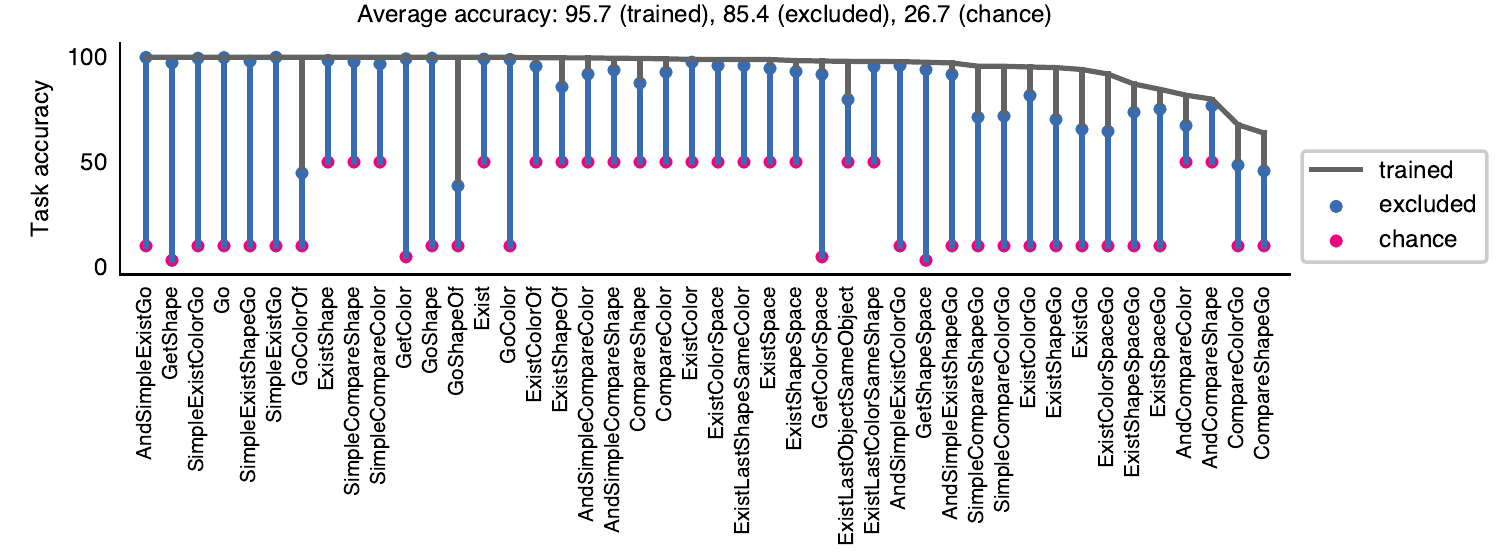}}
\caption{The proposed network can zero-shot generalize to new tasks. 44 networks were trained on 43 of 44 tasks. Shown are the maximum accuracies of the networks on the 43 trained tasks (gray), the one excluded (blue) task, and the chance levels for that task (red).}
\label{zeroshot}
\end{center}
\vskip -0.2in
\end{figure}

The most challenging form of generalization is to completely new tasks not explicitly trained on. To test whether the network can generalize to new tasks, we trained 44 groups of networks. Each group contains 10 networks and is trained on 43 out of 44 \COG tasks. We monitored the accuracy of all tasks. For each task, we report the highest accuracy across networks. We found that networks are able to immediately generalize to most untrained tasks (Figure \ref{zeroshot}). The average accuracy for tasks excluded during training ($85.4\%$) is substantially higher than the average chance level ($26.7\%$), although it is still lower than the average accuracy for trained tasks ($95.7\%$). Hence, our proposed model is able to perform zero-shot generalization across tasks with some success although not matching the performance as if trained on the task explicitly.

\subsection{Clustering and compositionality of the controller representation}
\begin{figure}[t]
\begin{center}
\centerline{\includegraphics[width=1.0\linewidth]{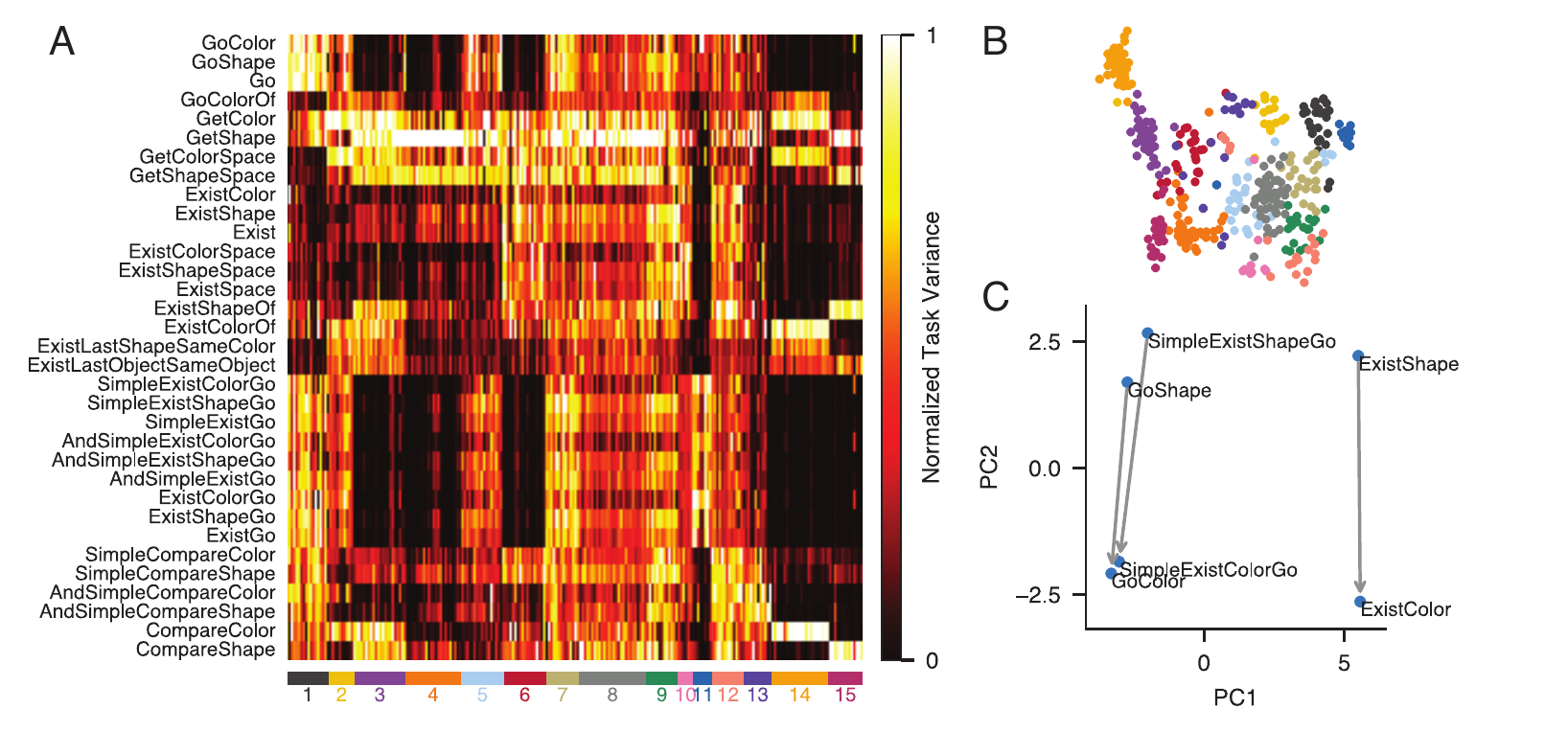}}
\caption{Clustering and compositionality in the controller. (\textbf{A}) The level of task involvement for each controller unit (columns) in each task (rows). The task involvement is measured by task variance, which quantifies the variance of activity across different inputs (task instructions and image sequences) for a given task. For each unit, task variances are normalized to a maximum of 1. Units are clustered (bottom color bar) according to task variance vectors (columns). Only showing tasks with accuracy higher than 90\%. (\textbf{B}) t-SNE visualization of task variance vectors for all units, colored by cluster identity. (\textbf{C}) Example compositional representation of tasks. We compute the state-space representation for each task as its mean controller activity vector, obtained by averaging across many different inputs for that task. The representation of 6 tasks are shown in the first two principal components. The vector in the direction of PC2 is a shared direction for altering a task to change from \textit{Shape} to \textit{Color}.}
\label{cluster_compositionality}
\end{center}
\vskip -0.2in
\end{figure}

To understand how the network is able to perform \COG tasks and generalize to new tasks, we carried out preliminary analyses studying the activity of the controller. One suggestion is that networks can perform many tasks by engaging clusters of units, where each cluster supports one operation \cite{yang2017clustering}. To address this question, we examined low-dimensional representations of the activation space of the controller and labeled such points based on the individual tasks. Figure \ref{cluster_compositionality}A and B highlight the clustering behavior across tasks that emerges from training on the \COG dataset (see Appendix for details).

Previous work has suggested that humans may flexibly perform new tasks by representing learned tasks in a compositional manner \cite{cole2013rapid,yang2017clustering}. For instance, the analysis of semantic embeddings indicates that network may learn shared directions for concepts across word embeddings \cite{mikolov2013distributed}.
We searched for signs of compositional behavior by exploring if directions in the activation space of the controller correspond to common sub-problems across tasks. Figure \ref{cluster_compositionality}C highlights a direction that was identified that corresponds to axis of {\tt Shape} to {\tt Color} across multiple tasks. These results provide a first step in understanding how neural networks can understand task structures and generalize to new tasks.

\section{Conclusions}
In this work, we built a synthetic, compositional dataset that requires a system to perform various tasks on sequences of images based on English instructions. The tasks included in our \COG dataset test a range of cognitive reasoning skills and, in particular, require explicit memory of past objects. This dataset is minimally-biased, highly configurable, and designed to produce a rich array of performance measures through a large number of named tasks.

We also built a recurrent neural network model that harnesses a number of attention and gating mechanisms to solve the \COG dataset in a natural, human-interpretable way. The model also achieves near state-of-the-art performance on another visual reasoning dataset, CLEVR. The model uses a recurrent controller to pay attention to different parts of images and instructions, and to produce verbal outputs, all in an iterative fashion. These iterative attention signals provide multiple windows into the model's step-by-step pondering process and provide clues as to how the model breaks complex instructions down into smaller computations. Finally, the network is able to generalize immediately to completely untrained tasks, demonstrating zero-shot learning of new tasks.


\newpage

\renewcommand\thesubsection{\Alph{subsection}}

\section*{Appendix}

\subsection{Operators and task graphs}
An operator is a simple function that receives and produces abstract data types such as an attribute, an object, a set of objects, a spatial range, or a Boolean. There are 8 operators in total: \textit{Select}, \textit{GetColor}, \textit{GetShape}, \textit{GetLoc}, \textit{Exist}, \textit{Equal}, \textit{And}, and \textit{Switch}.

The \textit{Select} operator is the most critical operator of all. It returns the set of objects that have certain attributes from a set of input objects. \textit{Select} can be instantiated with a color, a shape, a spatial range relative to a location, and a relative position in time (``now'', ``last'', ``latest''). By using ``last'' or ``latest'', a task can make inquiries about objects in the past, therefore demanding the network to have working memory. When the relative position in time is ``last'', the objects on the current image are not considered. Some instances of the \textit{Select} operator are \textit{Select}(ObjectSet, color=red, time=now), \textit{Select}(ObjectSet, shape=circle, time=last), \textit{Select}(ObjectSet, color=red, spatial range=left of (0.3, 0.8), time=latest). The attributes to be selected can also be outputs of other operators.

\textit{GetColor}, \textit{GetShape}, and \textit{GetLoc} returns the color, shape, and spatial location of an input object respectively. If the input is a set of object, and the set size is larger than 1, the output would be invalid, which would be propagated to the top of the graph. When the target response is invalid, no loss function is imposed for that image. When \textit{GetLoc} is used as the last operator of the graph, the task requires a pointing output.

\textit{Exist} returns a Boolean indicating whether the input set of objects is not empty. \textit{Equal} returns whether its two input attributes are the same. The input attributes can be color or shape. \textit{And} is the logical operator And. Finally, \textit{Switch} takes two operator subgraphs and a Boolean as inputs, returns the output of the first operator subgraph if the Boolean is True, and returns the output of the second subgraph otherwise. So the actual output of a \textit{Switch} operator can be either a pointing response or a verbal response.

Note that the simplicity of these operators is intuitive, but not rigorous. We chose operators that are relatively straightforward to humans. In contrast, for example, getting the quantitative area of an object would not be straightforward. The operators appear simple to the program because objects are already explicitly annotated with attributes such as colors and shapes.

The \COG dataset currently contains 44 tasks, with the number of operators in each task graph ranging from 2 to 11. Importantly, we consider the following four usages of the \textit{Select} operator and essentially treat them as separate operators: \textit{Select}(ObjectSet, color=X, time=T), \textit{Select}(ObjectSet, shape=X, time=T), \textit{Select}(ObjectSet, color=X, shape=Y, time=T), and \textit{Select}(ObjectSet, time=T), where T=now, last, latest. This means that we consider selecting the current red object and selecting the latest red object as different instances of the same task. But selecting the current red object and selecting the current circle would be considered instances of two different tasks.

Each task instruction is obtained from a task instance by traversing the task graph and combining pieces of text associated with each operator. For example, \textit{Select}(ObjectSet, shape=circle, color=red, time=now) is associated with "now red circle" and \textit{Exist}(X) is associated with "exist [text for X]". This method generates instructions that are often grammatically incorrect but still understandable to humans. However, this method can generate unnatural sentences when used on complicated task graphs, particularly when multiple \textit{Switch} are involved. In all of our tasks, at most one \textit{Switch} operator is involved.

\subsection{Minimizing response bias in the dataset}
When generating images for the \COG dataset, we start with target outputs that are minimally biased, then generate the images that would result in those outputs. To generate the images given the target outputs, we visit the sequence of images in the reverse chronological order -- the opposite direction of normal task execution. When visiting an image, we traverse the graph in a reverse topological order -- again, the opposite direction of normal task execution. When visiting each operator, we decide the supposed inputs to this operator given the target outputs, and the supposed inputs are typically passed on as target outputs of some other operators. Below we describe the supposed inputs given a target output for each operator. 

For \textit{Select}(ObjectSet, attribute=input attributes) and the target output, we will typically modify the set of object (ObjectSet) to sastisfy the target output. If the target output is a non-empty set of objects, then for each object in this output set, the ObjectSet should contain an object that satisfies both the input attributes being selected and the attributes of the output object. For example, if the operator is \textit{Select}(ObjectSet, color=red, time=now) and the target output is a single circle, then the ObjectSet must contain a red circle in the current image. We first search the ObjectSet to check if the appropriate object already exists. If so, nothing need to be done. If it does not exist, then we add one to the ObjectSet. When an attribute of the object to be added is not specified by either the input or the output, then it is randomly chosen from all the possible attribute values. When selecting an object using the temporal attribute ``last'' or ``latest'', we search $M_{\mathrm{max}}$ steps back in the history, excluding the current image for ``last''. If no satisfying object is found, we place one $M$ steps back, $M\sim U(0, M_{\mathrm{max}})$. This method ensures that the maximum memory duration for any object is $M_{\mathrm{max}}$. The expected memory duration would be $M_{\mathrm{max}}/3$. If the target output is an empty set, then we place a different object. We choose to place a different object here in order to prevent the network from solving some tasks by simply counting the number of objects. Furthermore, the object we place differs from the object to be selected by only one attribute. For example, if \textit{Select}(ObjectSet, color=red, shape=circle, time=now) has an empty target output, then we place either a red non-circle object or a non-red circle on the current image. If \textit{Select}(ObjectSet, spatial range=left of (0.5, 0.5)) has an empty target output, then we place an object at the right of (0.5, 0.5). This encourages the network to pay attention to all input attributes. 

For \textit{GetColor}, \textit{GetShape}, \textit{GetLoc}, the supposed input is a set of a single object with one attribute determined by the target output. For \textit{Exist}, the supposed input set of objects is non-empty if the target output is True, and empty if the target output is False. For \textit{Equal}(attribute1 , attribute2), we pass down two attributes that are either the same or different, based on the target output. For \textit{And}, both input Booleans will be True if the output is True. Otherwise, (Boolean1, Boolean2) would be (True, False), (False, True), (False, False) with probability $2\sqrt{0.5} - 1$, $2\sqrt{0.5} - 1$, $3 - 4\sqrt{0.5}$ respectively. These numbers are chosen such that Boolean1 and Boolean2 are statistically independent. \textit{Switch}(Boolean, operator1, operator2) does not support specification of a target output yet. Boolean is randomly chosen to be True or False.

\subsection{Canonical setting of the \COG dataset}
Here we explain the rationale for our choice of parameters for the canonical setting of \COG. We picked a small number of frames (4) for the canonical dataset because we needed to train thousands of networks. Having fewer frames greatly reduces the training time. We picked the highest possible value for memory duration, because time and memory are a major focus of the COG dataset. Finally, we picked the smallest non-zero value for the number of distractors because we wanted our model to reach high accuracy on the canonical dataset. The compositionality, interpretability, and zero-shot generalization analyses have little to tell if the full network is unable to learn the dataset well.

\subsection{\COG tasks}
\begin{figure*}[ht]
\begin{center}
\centerline{\includegraphics[width=\linewidth]{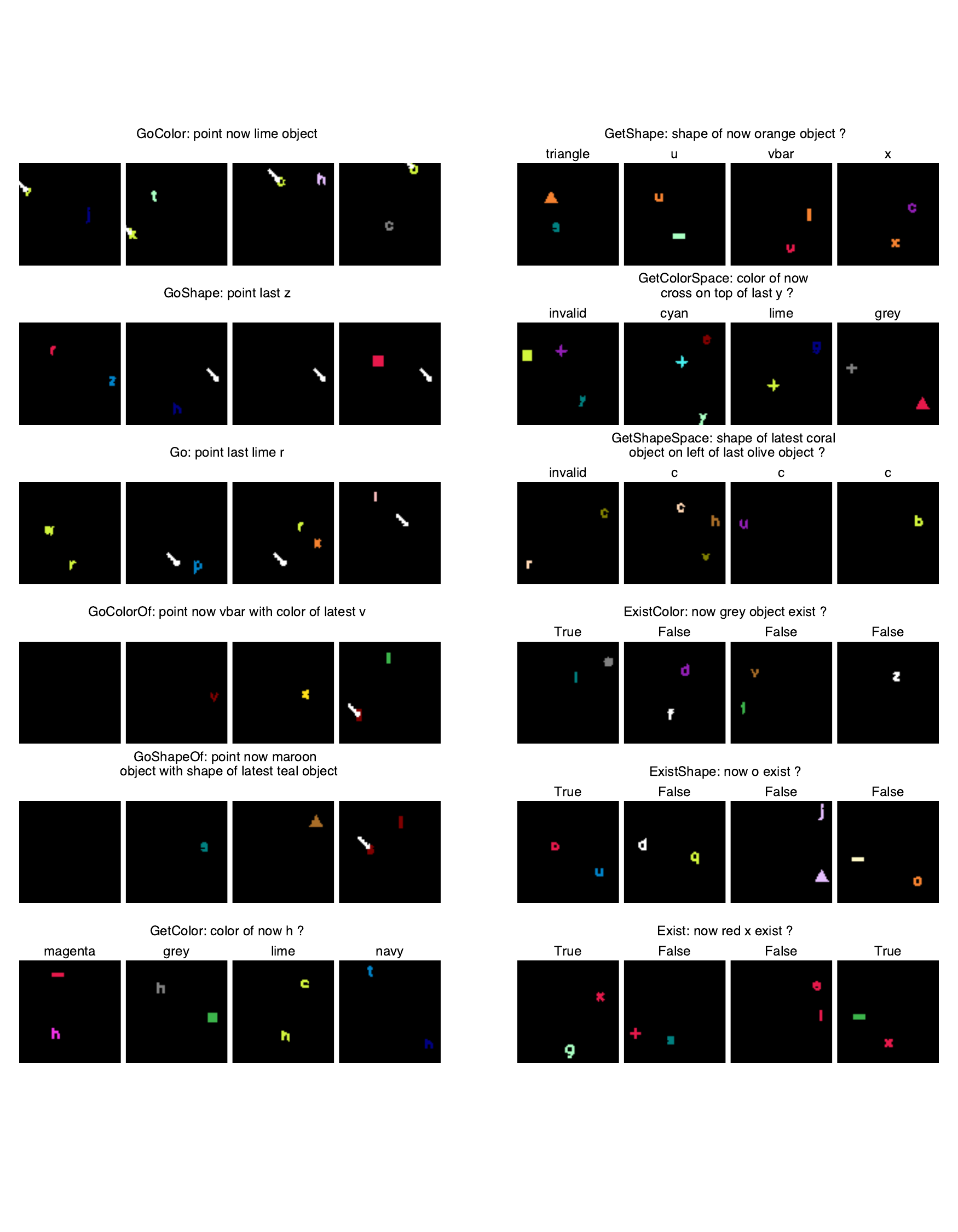}}
\caption{Example task instances for all tasks. Image resolution ($112 \times 112$) and task instructions are the same as shown to the network. White arrows indicate the target pointing output. No arrows are plotted if there is no valid target pointing output.}
\label{fig:all_task1}
\end{center}
\end{figure*}

\begin{figure*}[ht]
\begin{center}
\centerline{\includegraphics[width=\linewidth]{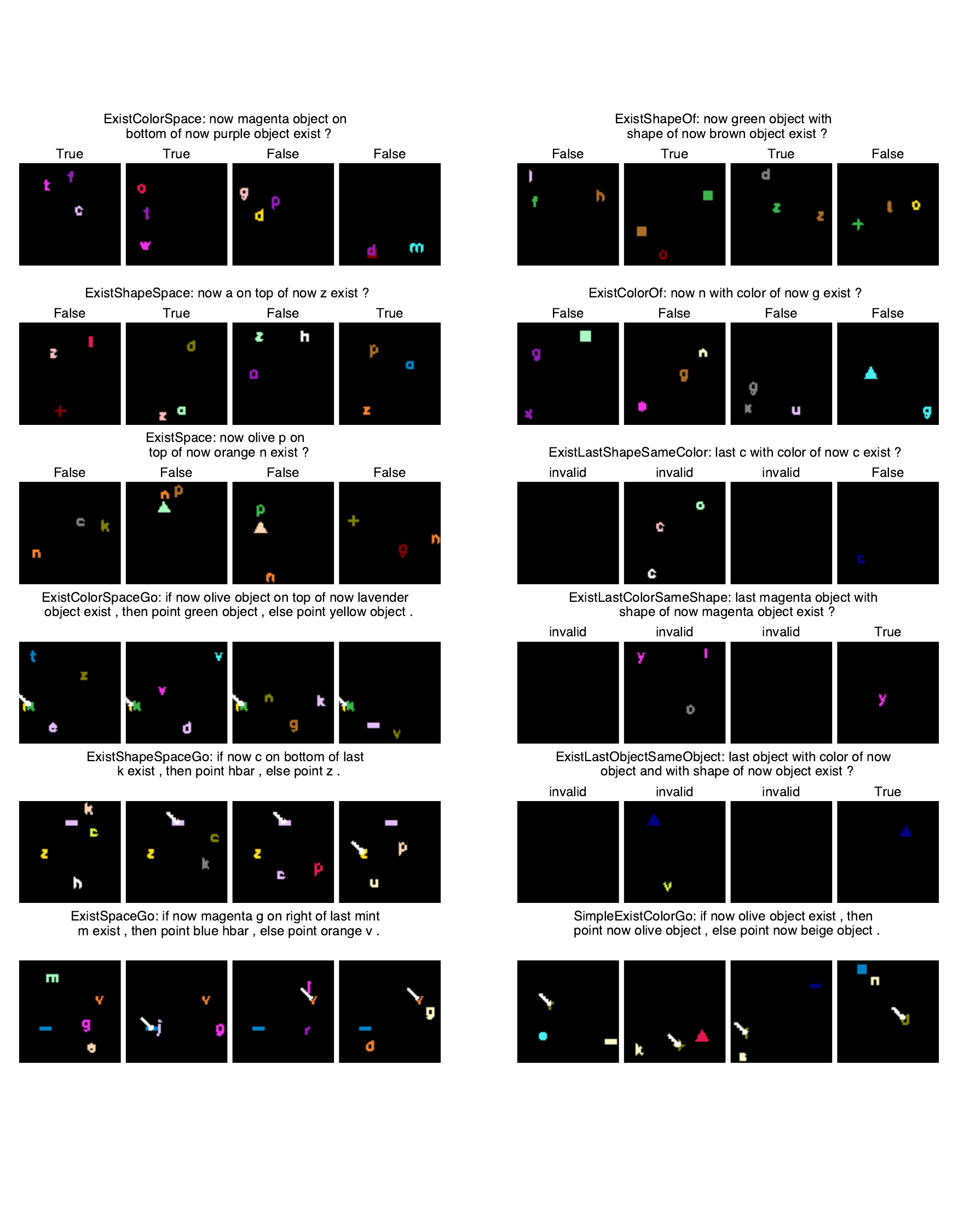}}
\caption{Example task instances for all tasks, continued.}
\label{fig:all_task2}
\end{center}
\end{figure*}

\begin{figure*}[ht]
\begin{center}
\centerline{\includegraphics[width=\linewidth]{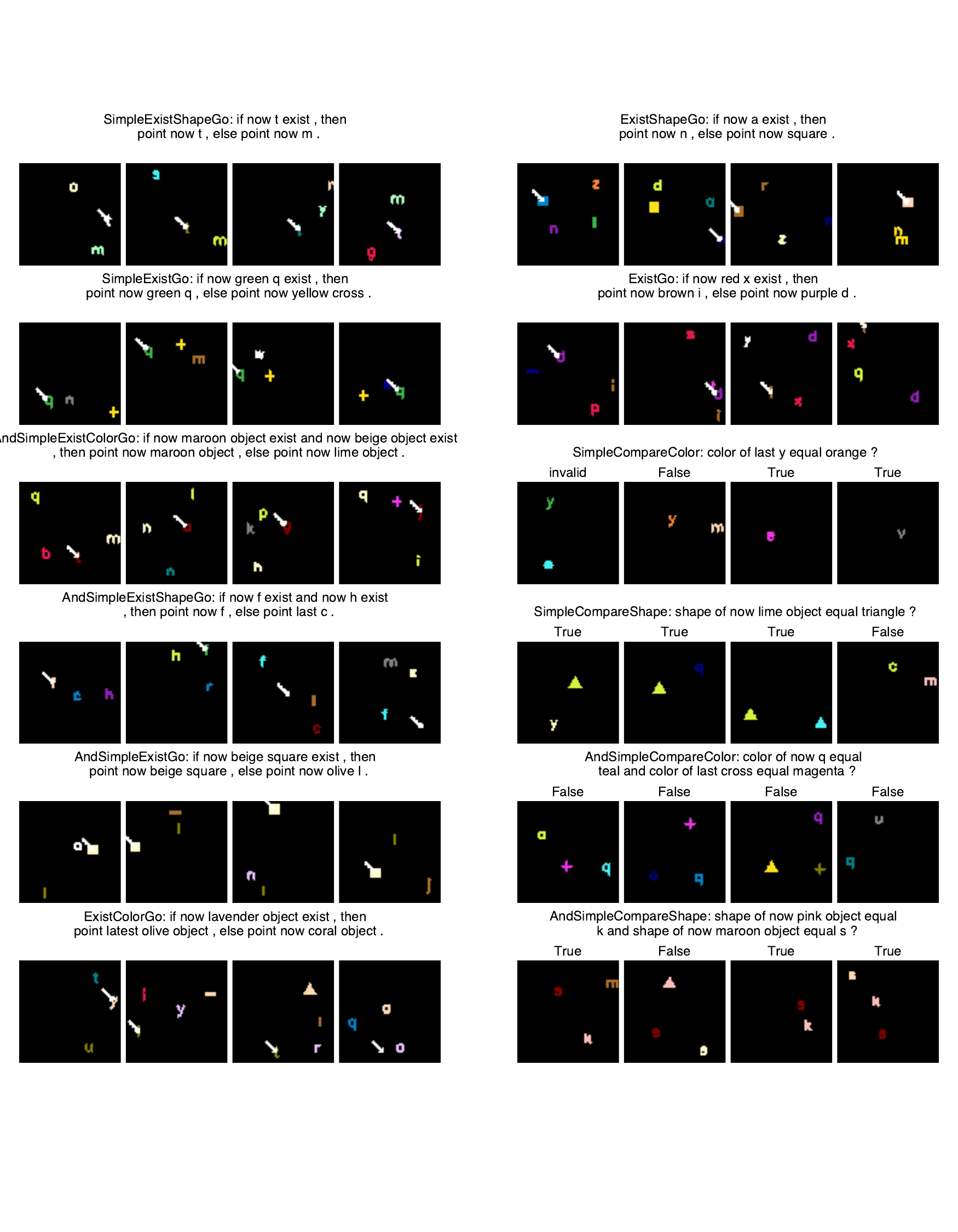}}
\caption{Example task instances for all tasks, continued.}
\label{fig:all_task3}
\end{center}
\end{figure*}

\begin{figure*}[ht]
\begin{center}
\centerline{\includegraphics[width=\linewidth]{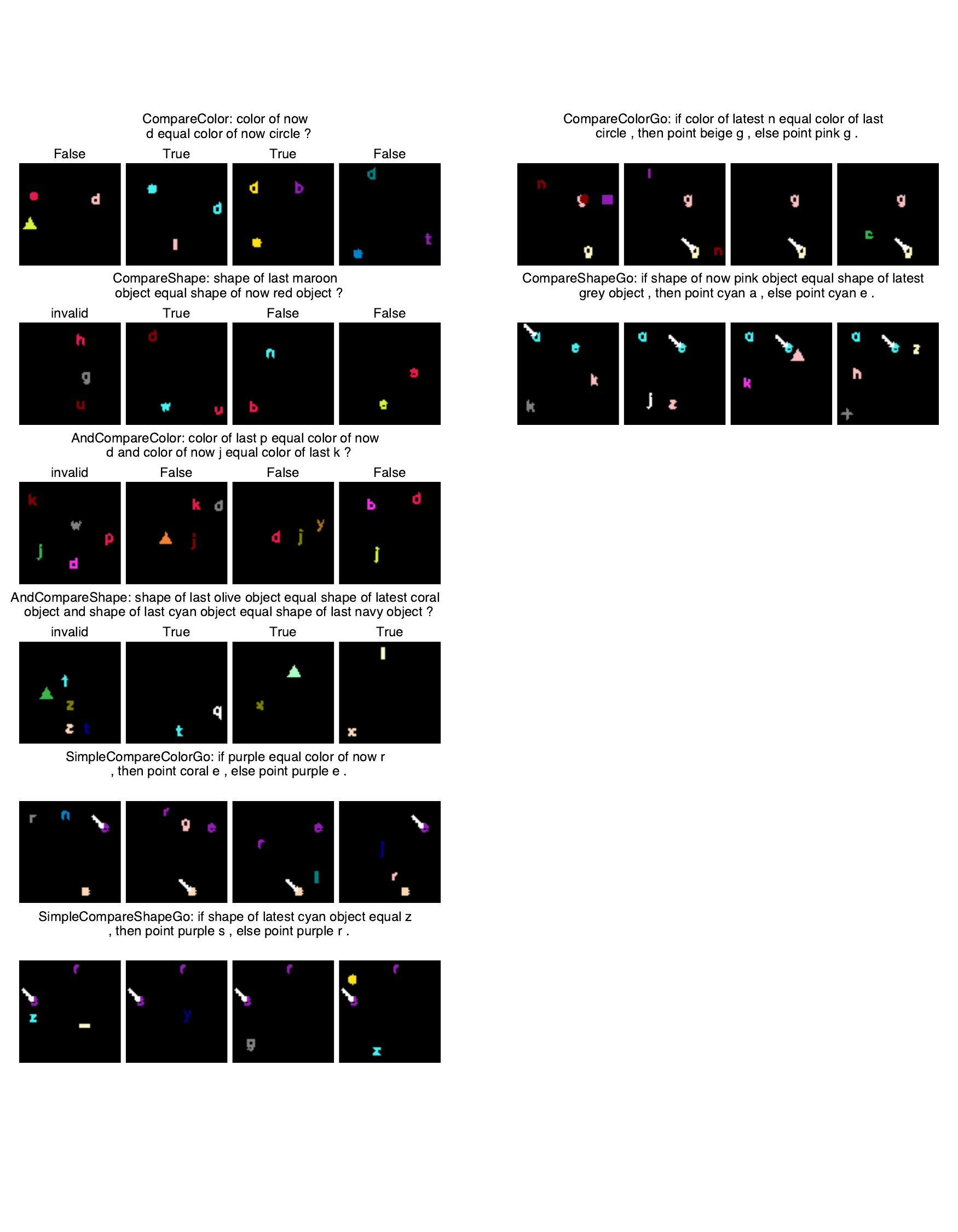}}
\caption{Example task instances for all tasks, continued.}
\label{fig:all_task4}
\end{center}
\end{figure*}

The \COG dataset contains 44 tasks. Of these 44 tasks, 39 tasks use the above method to generate its unbiased inputs. We include an additional 5 tasks that more directly mimic neuroscience and cognitive psychology experiments (e.g., delayed-match-to-sample and visual short-term-memory experiments), and we manually designed their input image sequences. These 5 tasks are \textit{GoColorOf}, \textit{GoShapeOf}, \textit{ExistLastShapeSameColor}, \textit{ExistLastColorSameShape}, and \textit{ExistLastObjectSameObject}. The number of distractors, expected memory duration, and number of effective images are fixed for these 5 tasks. In Figures \ref{fig:all_task1}-\ref{fig:all_task4}, we show example task instances for all tasks in the canonical \COG dataset.

\subsection{Training details}
All weights and biases are trained. No pretrained weights or embeddings are used. We use ReLU activations and Adam optimizer with default TensorFlow parameters and learning rate of $0.0005$. Training on CLEVR takes about 36 hours on a single Tesla K40 GPU. Training on canonical \COG takes about 34 hours on the same GPU. Hardest versions of COG take about twice as long to train. We use a batch size of $250$ for CLEVR. Each batch contains $25$ images with 10 questions per image. For \COG, we use a batch size of $48$. Each batch contains a random sample of task instances. Each task instance is generated from a randomly picked task, just in time for training. We train on CLEVR for about $107$ epochs and go over about $14.4$M task instances when training on \COG. We observe training instabilities about $10-15\%$ of the time when training versions of \COG with many frames, $F = 8$, and pondering steps, $6$. Testing on \COG was performed using $9600$ newly generated task instances for each task, giving a total of $422.4$k task instances.

\subsection{Detailed accuracy results on the CLEVR dataset}
We did not observe high variance on CLEVR during our work and did not run an experiment varying only the random variable initialization. As a related data point, the standard deviation of validation accuracies over a hyper-parameter tuning experiment with over 100 networks was 1.8\%. The actual standard deviation should be significantly smaller than 1.8\% since this experiment includes widely different hyper-parameter values in addition to random variable initialization.

\subsection{Potential strategies for performing standardized human experiments with the \COG dataset}
Obtaining human performance on the \COG dataset could help us better understand whether artificial neural networks solve cognitive tasks in ways similar to the human brain. However, to gain meaningful measurements of human performance on the \COG dataset, we need to specify several more parameters, each of which likely to have a major impact on the human performance measured. First, the time duration of each frame will be a critical parameter. Similar to the neural network presented in this work, having more time to process each frame will likely improve the performance of human subjects. Second, it is important to decide how the task instruction is presented to human subjects. If human subjects are allowed to be familiarized with the task instruction or the abstract task structure before viewing the visual stimuli, the performance will likely increase. Third, the size of the rendered images (or the distance between the monitor and the subject) could affect the performance. We do not have concrete suggestions on how to set these parameters yet.

\subsection{Tasks with more operators take longer to learn}
\begin{figure}[htbp]
\begin{center}
\centerline{\includegraphics[width=1.0\linewidth]{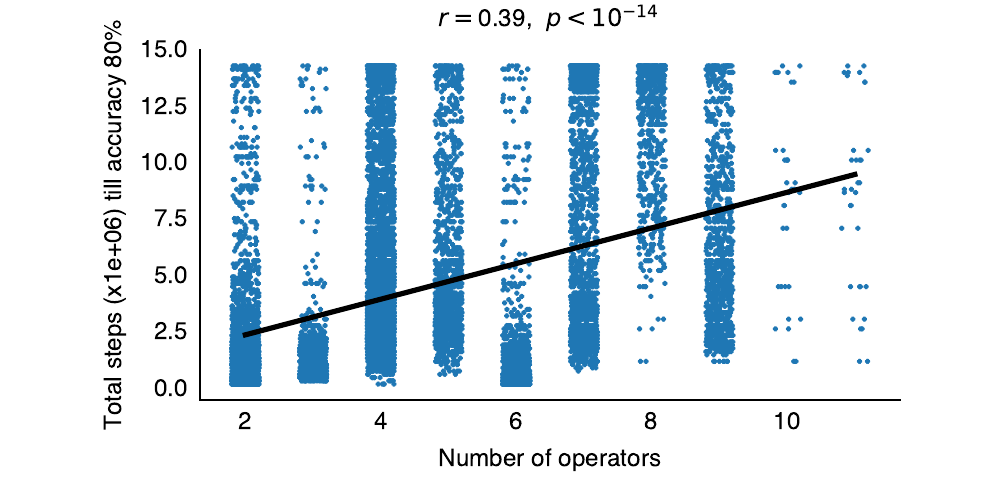}}
\caption{Tasks with more operators take longer to learn.} Each dot stands for the number of global training steps taken for a single task to reach $80\%$ accuracy in one of the 440 networks. Only showing results from tasks that reached $80\%$ accuracy.
\label{crossingtime_vs_numops}
\end{center}
\vskip -0.2in
\end{figure}

We found that tasks that contain more operators overall take longer to learn. We analyzed 440 networks trained for the zero-shot learning experiment, and found a highly significant correlation between the number of training steps required to reach $80\%$ accuracy and the number of operators in a task (Figure \ref{crossingtime_vs_numops}). However, the number of operators is by no means the only factor that affects the convergence time. Other factors including the average working memory load and the distance from other tasks can also have substantial impact the convergence time.

\subsection{Analyzing attention for the \COG dataset}
\begin{figure}[htbp]
\begin{center}
\centerline{\includegraphics[width=1.0\linewidth]{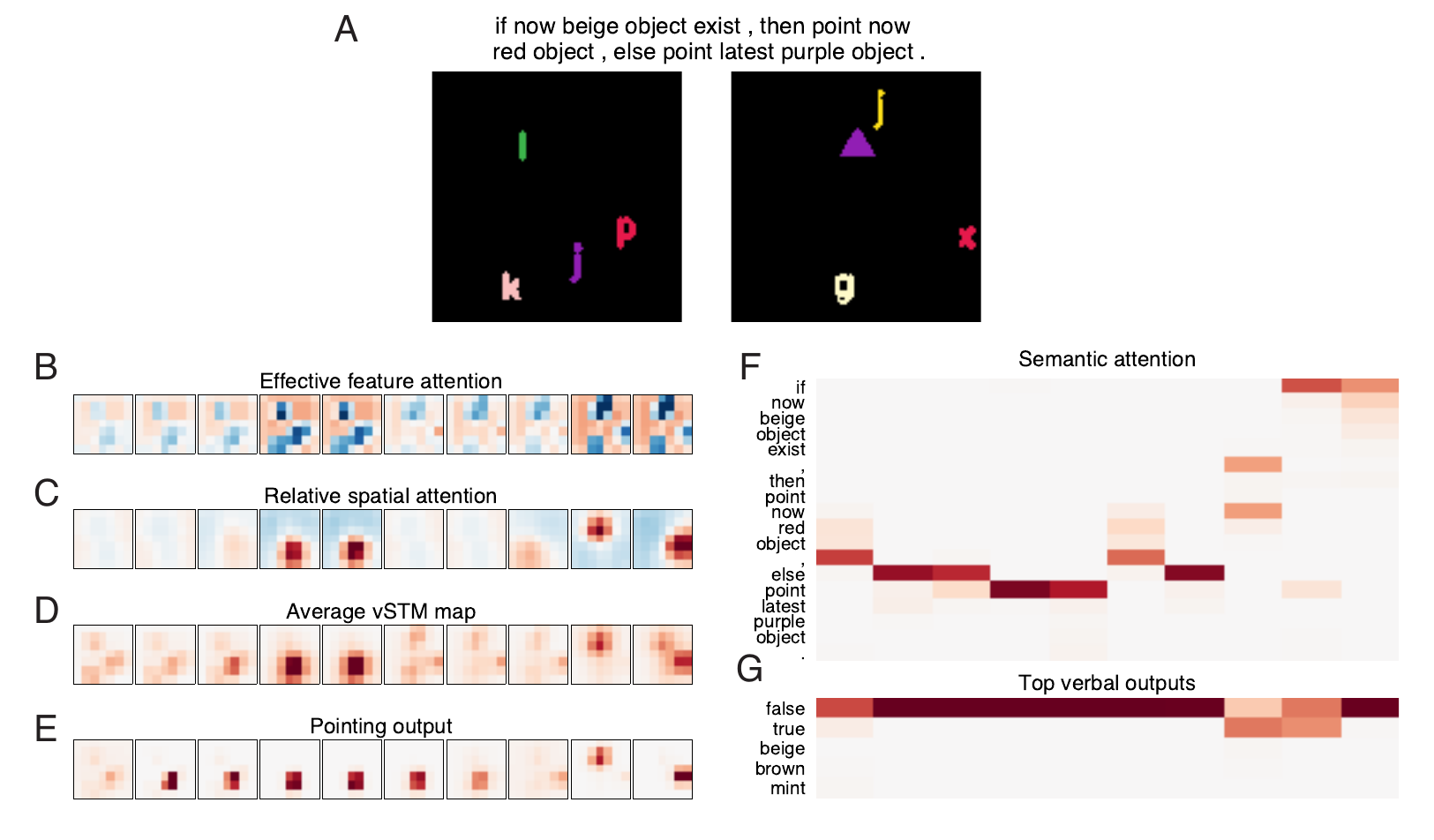}}
\caption{Visualization of network activity for single \COG example. (\textbf{A}) The task instruction and two example images shown sequentially to the network. (\textbf{B}) Effective feature attention. (\textbf{C}) Relative spatial attention. (\textbf{D}) Average vSTM map, computed by averaging the activity of all 4 vSTM maps. (\textbf{E}) Pointing output. (\textbf{F}) Semantic attention. (\textbf{G}) Top five verbal outputs during the network's pondering process. The network ponders for 5 steps for each image.}
\label{cog_attention}
\end{center}
\vskip -0.2in
\end{figure}

In Figure \ref{cog_attention}, we show a trained network solving an example from the \COG dataset. The network relies heavily on spatial attention, particularly late in the pondering process. It stores location information of objects in its vSTM maps even though that location information is not immediately used for generating the pointing response. 

\subsection{Task variance and compositionality}
The task variance $TV_{i,j}$ for controller unit $i$ and task $j$ is the variance of the unit's activity $r_{i,t}(u^{(j)})$ across all inputs (instructions and images) $u^{(j)}$ from task $j$, then averaged across pondering steps $t$. Mathematically,
\begin{equation*}
    TV_{i,j} = \langle \left[r_{i,t}(u^{(j)}) - \langle r_{i,t}(u^{(j)})\rangle_{u^{(j)}}\right]^2 \rangle_{u^{(j)}, t}.
\end{equation*}

The normalized task variance $\widehat{TV}_{i,j}$ is computed by normalizing the maximum task variance of any unit to 1.
\begin{equation*}
    \widehat{TV}_{i,j} = \frac{TV_{i,j}}{\max_j{TV_{i,j}}}.
\end{equation*}

The task variance vector for each unit $i$ is simply the vector formed by task variances for all tasks $j$. We exclude units with summed task variance less than 0.01, and exclude tasks with accuracy less than 90\% from our analysis. We ran 256 examples for each task to compute the task variance.

The task representation $\mathbf{v}$ used to show compositionality is computed by averaging the controller unit activity across tasks and pondering steps.
\begin{equation*}
    v_i = \langle r_{i,t}(u^{(j)}) \rangle_{u^{(j)}, j, t}.
\end{equation*}

\end{document}